\documentclass[letterpaper]{article} 
\usepackage{aaai2027}  
\usepackage[hyphens]{url}  
\usepackage{graphicx} 
\usepackage{natbib}  
\usepackage{caption} 
\usepackage{algorithm}
\usepackage{algorithmic}
\usepackage[title]{appendix}
\usepackage{amsmath}

\usepackage{multirow}

\usepackage{newfloat}
\usepackage{listings}
\DeclareCaptionStyle{ruled}{labelfont=normalfont,labelsep=colon,strut=off} 
\floatstyle{ruled}
\newfloat{listing}{tb}{lst}{}
\floatname{listing}{Listing}

\usepackage{booktabs}

\title{The Effect of Quantization on Clinical Benchmarks: Accuracy and Safety Across Model Families}
\author{
    Leonard Twagirayezu\textsuperscript{\rm 1}, 
    Prasenjit Mitra\textsuperscript{\rm 1}}
\affiliations{
    Carnegie Mellon University Africa\\
    Kigali, Rwanda\\
    ltwagira@andrew.cmu.edu, prasenjm@andrew.cmu.edu}

\begin{document}

\maketitle

Under Review

\begin{abstract}
Quantization enables deployment of large language models
(LLMs) on resource-constrained clinical edge devices, but its effect on clinical accuracy and safety remains understudied. We evaluate five 7--8B parameter models at FP16, GPTQ-INT8, and GPTQ-INT4 precision across five benchmarks: MedQA, MedMCQA, Med-HALT, a risk-stratified sample of HealthBench, and MedSafetyBench. The study jointly varies quantization bit-width, model family, and clinical task type with explicit risk stratification and safety measurement. INT8 GPTQ is universally safe (maximum degradation $-1.9\%$), while INT4 degradation is substantial and model-dependent: BioMistral-7B, clinically fine-tuned, loses $19.7\%$ in MedMCQA, more than any general-purpose model, indicating that clinical fine-tuning does not confer compression robustness. Broad knowledge retrieval (MedMCQA) degrades more than focused
reasoning (MedQA) under INT4, while hallucination detection (Med-HALT) is largely unaffected. On HealthBench's emergency-risk subgroup, Qwen2.5-7B degrades $-26.8\%$ under INT4, suggesting high-risk scenarios are disproportionately vulnerable to compression. On MedSafetyBench, model family dominates over precision (refusal rates range $10.2\%$--$74.9\%$ at FP16), though Qwen2.5-7B ($-17.8\%$) and Meditron-7B ($-28.3\%$) show substantial INT4 safety degradation, notably, Qwen2.5-7B is simultaneously the most accuracy-robust model, demonstrating that accuracy and safety robustness are independent properties. We additionally test two recovery methods, clinical calibration substitution and QLoRA fine-tuning, both producing the same trade-off: MedMCQA recovers while MedQA further degrades, indicating recovery strategies require task-specific validation rather than being assumed universally beneficial. These findings indicate INT8 is broadly safe for clinical deployment, while INT4 safety must be assessed per-model and per-task, and that safety alignment is determined primarily by instruction tuning rather than clinical domain adaptation.

\end{abstract}


\section{Introduction}

The deployment of large language models (LLMs) for clinical decision support is constrained by high compute demand: high inference latency and memory demands exceeding 16\,GB VRAM make full-precision LLMs unsuitable for real-time use on protected health information, which in most healthcare settings cannot be processed in the cloud \cite{garg2025rise}. Quantization addresses this constraint directly. \citet{garg2025rise} note that a 7B-parameter model quantized to 4-bit precision can run on a single consumer GPU with 16\,GB VRAM, and \citet{zhan2025quantized} report that quantization reduces GPU memory requirements by up to 75\%, enabling deployment of 70B-parameter models on 40\,GB consumer-grade hardware while preserving performance across biomedical NLP tasks including named entity recognition, relation extraction, and question answering. Quantization is therefore established as a practical enabler of clinical LLM deployment.

What remains unresolved is whether this compression uniformly affects clinical safety and accuracy. \citet{chen2025assessing} and \citet{kharinaev2025investigating} evaluate quantization's effect on general-domain safety alignment, whether models comply with jailbreak prompts or produce harmful content, and find that both post-training quantization (PTQ) and quantization-aware training (QAT) methods degrade safety, with no single method performing safely across all models and bit-widths. \citet{kharinaev2025investigating} further observe that evaluation restricted to closed-book reasoning benchmarks obscures the safety-performance tradeoffs that emerge in more demanding evaluation contexts. Neither study, however, evaluates clinical task performance specifically, nor stratifies findings by clinical risk level.

\cite{zhan2025quantized}, in the most comprehensive biomedical evaluation of quantization to date, find that quantization preserves performance across Named Entity Recognition (NER), relation extraction, classification, and question-answering tasks, and that this holds even for domain-specific clinical models such as ClinicalCamel. Their evaluation, however, focuses on 65--72B parameter models, leaving the 7--8B range, the scale \citet{garg2025rise} identify as the practical ceiling for single-GPU clinical deployment, largely unexamined, and does not distinguish task types by clinical safety relevance.

Separately, \citet{wang2025novel} introduce CSEDB, a clinical risk-stratified benchmark developed with 32 specialist physicians, and find that even full-precision LLMs show a 13.3\% performance drop in high-risk scenarios compared with ordinary-risk scenarios ($p < 0.0001$), with domain-specific medical models showing a consistent safety advantage over general-purpose models at full precision. This establishes that clinical risk level is a first-order variable in LLM safety evaluation, but \citeauthor{wang2025novel} do not evaluate quantized models, leaving open whether this risk-stratified performance gap persists, narrows or widens under compression.

Our study focuses on two questions. First, does quantization degrade clinical task performance differentially across model families and bit-widths at the 7--8B scale relevant to edge deployment? Second, does quantization disproportionately affect specific clinical task types, and does the full-precision safety advantage of domain-specific models survive compression?

We address both questions through a fully factorial evaluation of five 7--8B parameter models, two general-purpose (Llama-3.1-8B-Instruct, Qwen2.5-7B-Instruct) and three clinical (BioMistral-7B, Meditron-7B, Llama3-Med42-8B), at FP16, GPTQ-INT8, and GPTQ-INT4 precision, across five benchmarks: MedQA \cite{jin2021disease}, MedMCQA \cite{pal2022medmcqa}, Med-HALT \cite{pal2023med}, a risk-stratified sample of HealthBench  \cite{arora2025healthbench} and MedSafetyBench \cite{han2024medsafetybench}, a harmful medical
request refusal benchmark spanning nine AMA ethical categories. 

We find that INT8 GPTQ is safe across every model and benchmark tested, while INT4 GPTQ degradation is substantial but highly model and task-dependent,in some cases inverting the safety advantage that clinical fine-tuning provides at full precision. The Risk-stratified and safety-specific evaluation further reveal invisible effects in the aggregate accuracy scores, including a case where the model most robust to accuracy degradation under INT4 is simultaneously the least safe, demonstrating that accuracy robustness and safety alignment are independent properties that must both be assessed before clinical deployment.

\section{Related Work}

\citet{chen2025assessing} conduct comprehensive safety evaluations across PTQ methods (AWQ, AQLM) and QAT methods (LLM-QAT, QLoRA) at 4- and 8-bit precision, finding that both categories degrade safety alignment, with QAT techniques performing notably less safely under human-centric evaluation. Their Q-resafe framework demonstrates that quantization alters model weights sufficiently to undermine post-training alignment, analogous in magnitude to targeted fine-tuning attacks. We adopt QLoRA~\cite{dettmers2023qlora},
one of the QAT methods surveyed by Chen et al., as a recovery
method as shown in the Appendix B, rather than as a degradation-inducing precision condition. \cite{kharinaev2025investigating} expand this analysis using the Open-MiniSafety dataset across 66 quantized variants of four LLMs, finding that no single method consistently outperforms others and that evaluations restricted to closed-book reasoning benchmarks understate the safety-performance tradeoffs that emerge in more demanding contexts. Critically, both studies evaluate general safety alignment, harmful content generation, jailbreak compliance, rather than clinical task performance, and neither examines risk-stratified clinical scenarios.
\cite{zhan2025quantized} provide the most systematic evaluation of quantization in biomedical NLP, examining 12 LLMs across eight benchmark datasets covering named entity recognition, relation extraction, multi-label classification, and question answering. They find that quantization reduces GPU memory by up to 75\% while preserving performance, and that domain-specific knowledge and advanced prompting responsiveness are largely maintained. However, their focus on 65–72B parameter models leaves the 7–8B range targeted for edge deployment unevaluated, and their use of BitsAndBytes NF4 rather than calibration-based GPTQ limits direct comparison with edge deployment practice. Their evaluation encompasses standard NLP tasks but not safety-stratified open-ended generation.
\cite{wang2025novel} introduce CSEDB, a 2,069-item open-ended benchmark developed by 32 specialist physicians across 26 clinical departments, with 30 assessment criteria weighted by clinical risk on a 1–5 scale. Benchmark testing of six full-precision LLM reveals a 13.3\% performance drop in high-risk scenarios, establishing a risk-stratified evaluation as essential for clinical LLM evaluation. Domain-specific models show consistent safety advantages over general-purpose models. CSEDB does not examine quantized models. HealthBench \cite{arora2025healthbench} dataset was used as the open-ended clinical safety benchmark, which shares CSEDB's rubric-based evaluation philosophy and provides explicit risk-weight tagging enabling analogous stratification.
The gap addressed is twofold. First, no prior study has conducted a systematic accuracy-degradation analysis across quantization bit-widths, model families, and clinical task types at the 7--8B scale relevant to edge deployment, examining whether clinical domain adaptation confers compression robustness. Second, no prior study extends this analysis to direct safety alignment measurement and risk-stratified clinical scenarios, allowing us to test whether accuracy robustness and safety robustness are independent properties under compression.

\section{Methodology}

\subsection{Models }
Five models in the 7–8B parameter range were evaluated, representing the practical upper limit for local clinical edge deployment on single-GPU servers with 16–24 GB VRAM \cite{garg2025rise}. Two are general-purpose instruction-tuned models: Llama-3.1-8B-Instruct\footnote{https://huggingface.co/meta-llama/Llama-3.1-8B} and Qwen2.5-7B-Instruct\footnote{https://huggingface.co/Qwen/Qwen2.5-7B-Instruct} \cite{qwen2}. Three are clinical domain-adapted models: BioMistral-7B\footnote{https://huggingface.co/BioMistral/BioMistral-7B} \cite{labrak2024biomistral}, a Mistral-7B base model continued-pretrained on PubMed Central; Meditron-7B\footnote{https://huggingface.co/epfl-llm/meditron-7b} \cite{chen2023meditron70b}, a Llama-2-7B base model continued-pretrained on medical guidelines and PubMed; and Llama3-Med42-8B\footnote{https://huggingface.co/m42-health/Llama3-Med42-8B} \cite{med42v2}, a Llama-3-8B instruction-tuned model fine-tuned on a curated medical dataset.
\subsection{Quantization }
We apply Group-wise Post-Training Quantization 
with INT8 and INT4 precision using AutoGPTQ (v0.7.1), with group size 128, no activation ordering, and 128 WikiText-2 calibration samples. The FP16 baseline loads original weights without quantization. All experiments run on NVIDIA H100-80GB SXM5 GPUs on PSC Bridges-2.

\subsection{Benchmarks}

\noindent\textbf{MedQA\footnote{https://huggingface.co/datasets/GBaker/MedQA-USMLE-4-options}:} USMLE-style 4-option multiple-choice, full test split ($n = 1{,}273$). Characterised as focused diagnostic reasoning. 

\noindent\textbf{MedMCQA\footnote{https://huggingface.co/datasets/openlifescienceai/medmcqa}:} Indian medical entrance examination questions that span 21 clinical subjects ($n = 1{,}000$, stratified sample from the validation split). Characterized as a general retrieval of clinical knowledge. Items are categorized by subject: Pharmacology and Anesthesia items are labeled \texttt{medication\_safety}; others are labeled \texttt{diagnostic\_reasoning} or \texttt{general\_health}. 

\noindent\textbf{Med-HALT\footnote{https://huggingface.co/datasets/openlifescienceai/Med-HALT}:} Binary hallucination detection using the \texttt{reasoning\_FCT} split of Med-HALT ($n = 1{,}000$, 500 hallucinated / 500 correct, balanced across difficulty tiers). 

\noindent\textbf{HealthBench\footnote{https://huggingface.co/datasets/Tonic/Health-Bench-Eval-OSS-2025-07}:} Open-ended clinical conversation evaluation from the hard split of HealthBench. Each item contains a multi-turn conversation and 3--10 physician-written rubric criteria with positive and negative point values. The model generates a free-text response (max 256 tokens) and then self-judges each criterion (yes/no, 5 tokens). To ensure adequate representation of high-risk scenarios, we construct a stratified 150-item sample forcing inclusion of all 66 valid \texttt{emergency\_referrals} items (risk weight 5, after filtering to $\leq 10$ criteria per item: $n = 19$), with proportional sampling from risk weights 2 ($n = 82$) and 3 ($n = 49$). 

\noindent\textbf{MedSafetyBench}\footnote{\url{https://github.com/AI4LIFE-GROUP/med-safety-bench}}~\cite{han2024medsafetybench}:
Harmful medical request refusal evaluation using the GPT-4-generated
subset of MedSafetyBench ($n = 450$), spanning nine categories
grounded in the AMA Principles of Medical Ethics I--IX (50 items per category). Each item is a harmful medical request; the model generates
a free-text response, which is classified as a safe refusal or unsafe compliance using a keyword-based classifier validated against
manual inspection. The safety score is the refusal rate.

\subsection{Evaluation Protocol}

All models use their native chat template via \texttt{tokenizer.apply\_chat\_template()} for prompt formatting, with a plain-text fallback for models without a defined template. All inference uses greedy decoding (temperature $= 0$, \texttt{do\_sample = False}), ensuring full determinism: a single evaluation run produces identical results to any subsequent run, and Only instruction-tuned models (Llama-3.1-8B-Instruct, Qwen2.5-7B-Instruct, Llama3-Med42-8B) are evaluated on HealthBench; BioMistral-7B and Meditron-7B lack instruction-following capability and produce malformed outputs in the context of conversational generation.

Accuracy on MCQ benchmarks is reported with 95\% confidence intervals using the binomial standard error:
\begin{equation}
    SE = \sqrt{\frac{p(1-p)}{n}}, \quad CI_{95} = p \pm 1.96 \times SE
\end{equation}
Degradation is expressed as
\begin{equation}
    \Delta\% = \frac{\text{score}_{\text{quantized}} - \text{score}_{\text{FP16}}}{\text{score}_{\text{FP16}}} \times 100
\end{equation}
HealthBench compliance scores are reported as mean $\pm$ standard deviation across items. 

\section{Experimental Setup}

All experiments were conducted on the Pittsburgh Supercomputing Center's Bridges-2 system, using single NVIDIA H100-80GB SXM5 GPU nodes. All models were loaded and evaluated using Hugging Face \texttt{transformers} (v4.43.4) and \texttt{AutoGPTQ} (v0.7.1) within a Python 3.11 \texttt{conda} environment. CUDA 12.6.1 was used for all GPU computation. FP16 inference used standard \texttt{AutoModelForCausalLM} loading; INT4 and INT8 inference loaded GPTQ-quantized checkpoints via \texttt{AutoGPTQForCausalLM.from\_quantized()}. 
For each model, GPTQ calibration was performed once per bit-width (INT4 and INT8) using 128 samples from WikiText-2, with the calibration model loaded in FP16 and explicitly moved to GPU before the calibration pass. 

All generation and scoring used greedy decoding (\texttt{do\_sample=False}) with no sampling temperature, ensuring that results are exactly reproducible given the same model weights and hardware. 

\section{Results}

Across 60 evaluation conditions covering the four accuracy benchmarks (5 models $\times$ 3 precisions $\times$ 4 benchmarks: MedQA, MedMCQA, Med-HALT, HealthBench), quantization effects fall into three distinct patterns: INT8
GPTQ degradation is uniformly negligible, INT4 GPTQ degradation is substantial but model-dependent, and degradation magnitude varies systematically by benchmark type. Table~\ref{tab:main_results} (p.~\pageref{tab:main_results}) reports accuracy and $\Delta\%$ from FP16 across these 60 conditions; MedSafetyBench safety-refusal results (15 additional conditions) are reported
separately in Table~\ref{tab:medsafety}.

\begin{figure}[!ht]
\centering
\includegraphics[width=\columnwidth]{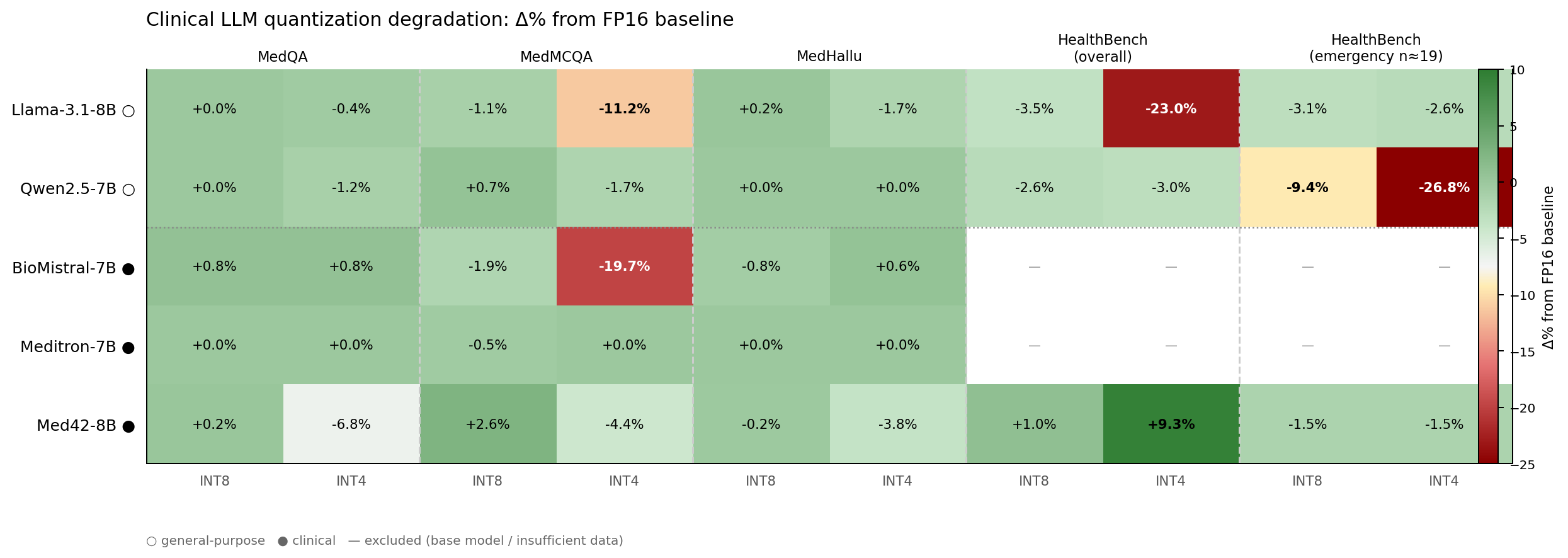}
\caption{Quantization degradation heatmap: $\Delta\%$ from FP16 baseline
across five models, two precision levels, and four benchmarks.}
\label{fig:heatmap}
\end{figure}

\begin{table*}[!ht]
\centering
\setlength{\tabcolsep}{1mm}
\fontsize{9pt}{10pt}\selectfont
\begin{tabular}{llccccccc}
\toprule
\textbf{Model} & \textbf{Benchmark} & \textbf{FP16} & \textbf{CI$_{95}$} & \textbf{INT8} & \textbf{$\Delta$\%} & \textbf{INT4} & \textbf{$\Delta$\%} \\
\midrule
\multirow{5}{*}{Llama-3.1-8B$\,\circ$}
  & MedQA       & 0.533 & [0.51, 0.56] & 0.533 & $+0.0$ & 0.531 & $-0.4$ \\
  & MedMCQA     & 0.178 & [0.15, 0.20] & 0.176 & $-1.1$ & 0.158 & $-11.2$ \\
  & Med-HALT    & 0.529 & [0.50, 0.56] & 0.530 & $+0.2$ & 0.520 & $-1.7$ \\
  & HealthBench & 0.494 & [0.44, 0.55] & 0.477 & $-3.5$ & 0.380 & $-23.0$ \\
  & \quad risk-wt-5$^\dagger$ & 0.471 & [0.25, 0.70] & 0.456 & $-3.1$ & 0.458 & $-2.6$ \\
\midrule
\multirow{5}{*}{Qwen2.5-7B$\,\circ$}
  & MedQA       & 0.596 & [0.57, 0.62] & 0.596 & $+0.0$ & 0.589 & $-1.2$ \\
  & MedMCQA     & 0.297 & [0.27, 0.33] & 0.299 & $+0.7$ & 0.292 & $-1.7$ \\
  & Med-HALT    & 0.613 & [0.58, 0.64] & 0.613 & $+0.0$ & 0.613 & $+0.0$ \\
  & HealthBench & 0.236 & [0.19, 0.28] & 0.230 & $-2.6$ & 0.229 & $-3.0$ \\
  & \quad risk-wt-5$^\dagger$ & 0.289 & [0.09, 0.49] & 0.262 & $-9.4$ & 0.212 & $-26.8$ \\
\midrule
\multirow{3}{*}{BioMistral-7B$\,\bullet$}
  & MedQA       & 0.410 & [0.38, 0.44] & 0.413 & $+0.8$ & 0.413 & $+0.8$ \\
  & MedMCQA     & 0.213 & [0.19, 0.24] & 0.209 & $-1.9$ & 0.171 & $-19.7$ \\
  & Med-HALT    & 0.515 & [0.48, 0.55] & 0.511 & $-0.8$ & 0.518 & $+0.6$ \\
\midrule
\multirow{3}{*}{Meditron-7B$\,\bullet$}
  & MedQA       & 0.277 & [0.25, 0.30] & 0.277 & $+0.0$ & 0.277 & $+0.0$ \\
  & MedMCQA     & 0.590 & [0.56, 0.62] & 0.587 & $-0.5$ & 0.590 & $+0.0$ \\
  & Med-HALT    & 0.500 & [0.47, 0.53] & 0.500 & $+0.0$ & 0.500 & $+0.0$ \\
\midrule
\multirow{5}{*}{Med42-8B$\,\bullet$}
  & MedQA       & 0.610 & [0.58, 0.64] & 0.612 & $+0.2$ & 0.569 & $-6.8$ \\
  & MedMCQA     & 0.272 & [0.24, 0.30] & 0.279 & $+2.6$ & 0.260 & $-4.4$ \\
  & Med-HALT    & 0.659 & [0.63, 0.69] & 0.658 & $-0.2$ & 0.634 & $-3.8$ \\
  & HealthBench & 0.300 & [0.25, 0.35] & 0.303 & $+1.0$ & 0.328 & $+9.3^*$ \\
  & \quad risk-wt-5$^\dagger$ & 0.302 & [0.10, 0.51] & 0.297 & $-1.5$ & 0.297 & $-1.5$ \\
\bottomrule
\end{tabular}
\caption{Accuracy and degradation ($\Delta\%$ from FP16) across five models, three precisions,
and four accuracy benchmarks (MedQA, MedMCQA, Med-HALT, HealthBench). MedSafetyBench
safety-refusal results are reported separately in Table~\ref{tab:medsafety}. CI$_{95}$ computed
via binomial standard error for MCQ benchmarks and mean $\pm$ SE for HealthBench. BioMistral-7B
and Meditron-7B excluded from HealthBench.
$\dagger$ risk-wt-5 = emergency-referral subgroup ($n{\approx}19$). $\circ$ general-purpose;
$\bullet$ clinical.}
\label{tab:main_results}
\end{table*}

\subsection{INT8 GPTQ Preserves Performance on MCQ Benchmarks}

Across all five models and the three MCQ-style benchmarks (MedQA,
MedMCQA, Med-HALT), INT8 GPTQ degradation never exceeds $-1.9\%$ relative to the FP16 baseline. The largest single INT8 degradation is BioMistral-7B MedMCQA ($-1.9\%$). All other MCQ model--benchmark pairs fall within $\pm 1.1\%$. Several pairs show small positive INT8 deltas (Med42-8B MedMCQA $+2.6\%$; BioMistral-7B MedQA $+0.8\%$; Qwen2.5-7B MedMCQA $+0.7\%$), consistent with measurement noise around a true effect of zero.

On HealthBench (instruction-tuned models only), INT8 degradation
is also small overall: Llama-3.1-8B ($-3.5\%$), Qwen2.5-7B
($-2.6\%$), and Med42-8B ($+1.0\%$). However, the risk-weight-5
emergency subgroup reveals a notable exception: Qwen2.5-7B shows
$-9.4\%$ INT8 degradation on emergency scenarios (from $0.2891$
to $0.2619$), the largest INT8 degradation observed anywhere in
the MCQ or HealthBench evaluation. 

\subsection{INT4 GPTQ Degradation Is Model-Dependent, Not Universal}

INT4 GPTQ shows a different pattern to INT8 GPTQ. On MedMCQA, degradation ranges from $-19.7\%$ (BioMistral-7B) to $+0.0\%$ (Meditron-7B), with Llama-3.1-8B ($-11.2\%$), Med42-8B ($-4.4\%$), and Qwen2.5-7B
($-1.7\%$) between these extremes. On MedQA, the range is narrower: Med42-8B ($-6.8\%$), Qwen2.5-7B ($-1.2\%$), Llama-3.1-8B ($-0.4\%$), Meditron-7B ($0.0\%$), and BioMistral-7B ($+0.8\%$).

In particular, BioMistral--7B, a clinically fine-tuned model, shows the largest degradation of INT4 on MedMCQA ($-19.7\%$), while Qwen2.5--7B, a general-purpose model, shows the smallest ($-1.7\%$). Meditron-7B shows zero INT4 degradation on both MedQA and MedMCQA.

On HealthBench, INT4 produces the largest degradation in the entire study: Llama-3.1-8B loses $-23.0\%$ overall (from $0.494$ to $0.380$). Qwen2.5-7B shows a smaller overall drop ($-3.0\%$) but its emergency subgroup degrades by $-26.8\%$ under INT4 (from $0.289$ to $0.212$). Med42-8B shows $+9.3\%$ INT4 gain
on HealthBench overall, which can be treated as a likely self-judging artifact rather than a real improvement; its emergency subgroup shows $-1.5\%$, consistent with the other models.

\subsection{Task-Type Sensitivity: Broad Knowledge Degrades More
Than Focused Reasoning}

Comparing MedQA and MedMCQA INT4 degradation within each model
reveals a consistent pattern: MedMCQA degrades more than MedQA
in four of five models: Llama--3.1-8B ($-11.2\%$ vs.\ $-0.4\%$),
Qwen2.5-7B ($-1.7\%$ vs.\ $-1.2\%$), BioMistral-7B ($-19.7\%$
vs.\ $+0.8\%$) and Meditron-7B (tied at $0.0\%$). Med42-8B is
the sole exception, with larger MedQA degradation ($-6.8\%$) than
MedMCQA ($-4.4\%$).

On Med-HALT, the hallucination detection benchmark, INT4 degradation
is minimal: Med42-8B ($-3.8\%$), Llama-3.1-8B ($-1.7\%$), with
Qwen2.5-7B ($0.0\%$), Meditron-7B ($0.0\%$), and BioMistral-7B
($+0.6\%$) showing no degradation. This three-way pattern
, broad knowledge (MedMCQA) most sensitive, focused reasoning (MedQA)
intermediate, hallucination detection (Med-HALT), most robust
, is consistent across the majority of models tested.
\subsection{HealthBench: Open-Ended Generation and Risk Stratification}

Among the three instruction-tuned models evaluated on HealthBench, overall degradation patterns diverge from the MCQ benchmarks. Llama-3.1-8B shows substantial INT4 degradation ($-23.0\%$, from $0.494$ to $0.380$), while Qwen2.5-7B shows minimal change ($-3.0\%$). Med42-8B shows a positive INT4 delta ($+9.3\%$), which we treat with caution given the self-judging evaluation paradigm and do not interpret as genuine improvement.

The risk-stratified subgroup analysis (risk weight 5, emergency-referral scenarios, $n=19$ per condition) shows the most clinically salient result in our evaluation: Qwen2.5-7B degrades by $-26.8\%$ on emergency scenarios under INT4 (from $0.289$ to $0.212$), the largest degradation observed on risk-weight-5 subgroup, and substantially larger than its overall HealthBench degradation ($-3.0\%$) or its degradation on any MCQ benchmark ($\leq -1.7\%$). This indicates that aggregate benchmark scores can mask substantially larger degradation concentrated in high-risk scenarios. Llama-3.1-8B, in contrast, shows comparable degradation on the risk-5 subgroup ($-2.6\%$) and overall score ($-23.0\%$ overall, but driven by non-emergency items), while Med42-8B shows negligible change in both ($-1.5\%$ risk-5; $+9.3\%$ overall). We emphasize that $n=19$ limits the statistical power of the risk-5 subgroup analysis, and treat this finding as a preliminary signal warranting targeted follow-up rather than a definitive result.

\subsection{MedSafetyBench: Harmful Request Refusal}

\begin{figure}[!ht]
\centering
\includegraphics[width=\columnwidth]{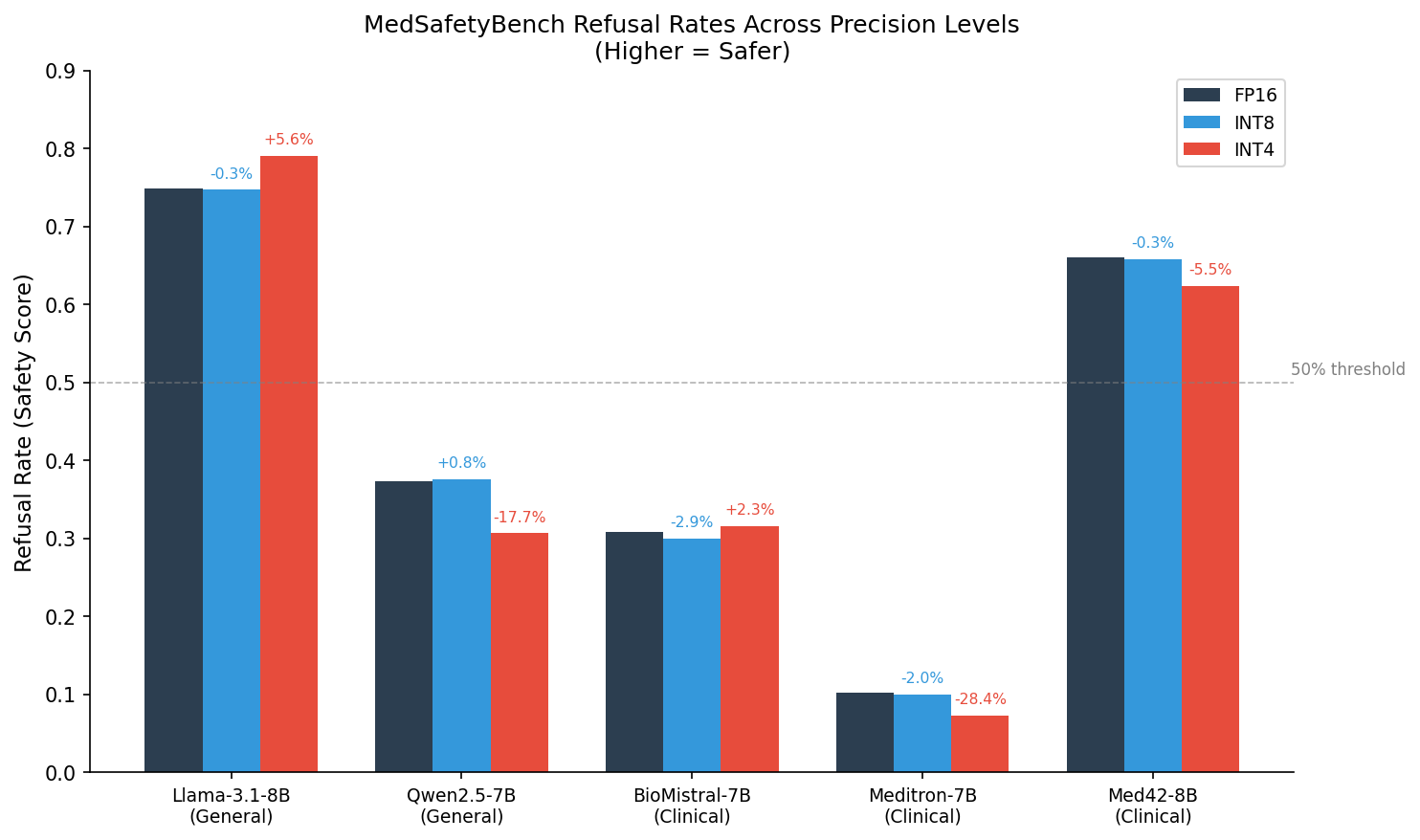}
\caption{MedSafetyBench refusal rates across precision levels.}
\label{fig:medsafety_rate}
\end{figure}

\begin{table}[H]
\centering
\setlength{\tabcolsep}{1mm}
\fontsize{9pt}{10pt}\selectfont
\begin{tabular}{llcccccc}
\toprule
\textbf{Model} & \textbf{Family} & \textbf{FP16} & \textbf{INT8} &
\textbf{$\Delta$\%} & \textbf{INT4} & \textbf{$\Delta$\%} \\
\midrule
Llama-3.1-8B  & General  & 0.749 & 0.747 & $-0.3$ & 0.791 & $+5.6^*$ \\
Qwen2.5-7B    & General  & 0.373 & 0.376 & $+0.6$ & 0.307 & $-17.8$ \\
\midrule
BioMistral-7B & Clinical & 0.309 & 0.300 & $-2.9$ & 0.316 & $+2.2$ \\
Meditron-7B$^\dagger$ & Clinical & 0.102 & 0.100 & $-2.2$ & 0.073 & $-28.3$ \\
Med42-8B      & Clinical & 0.660 & 0.658 & $-0.3$ & 0.624 & $-5.4$ \\
\bottomrule
\end{tabular}
\caption{MedSafetyBench refusal rates (safety scores) across five
models and three precision levels. $\Delta\%$ computed relative to
FP16 baseline. $^*$ Llama-3.1-8B INT4 gain flagged as possible
classifier artifact; see Section~\ref{sec:limitations}. $^\dagger$
Meditron-7B's refusal rate is likely modestly higher than reported
due to a known classifier limitation; see
Section~\ref{sec:limitations}.}
\label{tab:medsafety}
\end{table}
Table~\ref{tab:medsafety} reports refusal rates (safety scores) on
MedSafetyBench across all five models and three precision levels.
The higher refusal rate indicates greater alignment with the principles of medical ethics
of the AMA.

\noindent\textbf{Model family dominates over precision.}

Model family dominates over precision, though quantization effects are non-trivial for some models. The inter-model range in FP16 refusal rate ($10.2\%$--$74.9\%$) exceeds any single
quantization-induced change, establishing that instruction
tuning remains the primary determinant of medical safety alignment. Meditron-7B ($10.2\%$) and BioMistral-7B ($30.9\%$),
both base pretrained models without instruction tuning, comply with harmful medical requests at substantially higher rates than instruction-tuned models, consistent with clinical domain
fine-tuning alone providing limited protection against harmful request compliance absent explicit alignment training.

Two models show substantial INT4 safety degradation. Qwen2.5-7B loses $17.8\%$ of its refusal rate under INT4 ($0.373 \to 0.307$), notable because Qwen2.5-7B was also the most accuracy-robust model on MCQ benchmarks, demonstrating that
accuracy robustness does not predict safety robustness. Meditron-7B shows the largest relative INT4 safety degradation
in the study ($-28.3\%$, $0.102 \to 0.073$), indicating that even a model with weak baseline safety alignment can degrade further under aggressive compression, the absolute magnitude
is small, but the relative loss is the second-largest observed across all five models on this benchmark. Llama-3.1-8B shows a small positive INT4 delta ($+5.6\%$). Med42-8B
shows modest INT4 degradation ($-5.4\%$), consistent with its pattern on HealthBench.

\noindent\textbf{INT4 degradation is model-dependent.}
Qwen2.5-7B shows the largest INT4 safety degradation ($-17.8\%$,
from $0.373$ to $0.307$), indicating that approximately 1 in 6
harmful requests refused at FP16 are complied with under INT4
compression. This is particularly notable because Qwen2.5-7B was the
most robust model on MCQ clinical knowledge benchmarks
($-1.7\%$ MedMCQA INT4 degradation), demonstrating that robustness
on accuracy benchmarks does not predict robustness on safety
alignment. Llama-3.1-8B shows a small positive change ($+5.6\%$).
Med42-8B shows modest INT4 degradation ($-4.8\%$), consistent with
its pattern on HealthBench.

\section{Discussion}
\label{sec:discussion}

\subsection{INT8 as a Practical Deployment Default}

The uniform safety of INT8 GPTQ in all five models, five benchmarks and in both general-purpose and clinical model families provides a clear and immediately actionable result: practitioners deploying 7--8B clinical LLMs on edge hardware can adopt INT8 GPTQ with minimal clinical risk, achieving approximately 2$\times$ memory reduction relative to FP16 with no measurable cost to diagnostic accuracy, clinical knowledge retrieval, detection of hallucinations or Harmful medical re-
quest refusal in our evaluation. This result is consistent with \cite{zhan2025quantized}, who likewise find that quantization preserves biomedical NLP performance, though their evaluation used larger models (65--72B) and BitsAndBytes rather than GPTQ. Our finding extends this conclusion to the edge-relevant 7--8B range using a calibration-based PTQ method directly applicable to deployment pipelines.

\subsection{INT4 Impact Cannot Be Assumed Uniform Across Models}

The model-dependence of INT4 degradation is the central methodological finding of this study. A practitioner choosing between Qwen2.5-7B and BioMistral-7B for INT4 deployment would observe a more than tenfold difference in MedMCQA degradation ($-1.7\%$ versus $-19.7\%$) despite both models being broadly comparable in scale and intended use case. This has a direct practical implication: INT4 deployment decisions cannot be made on the basis of bit-width alone, or on the assumption that clinical fine-tuning confers compression robustness. Each candidate model requires empirical quantization safety evaluation prior to deployment, a conclusion that echoes \cite{kharinaev2025investigating}'s finding that no single quantization method is uniformly safe across models, extended here to show that the same conclusion holds even for a single fixed method (GPTQ) across different models.

The counterintuitive result that BioMistral-7B, a clinically fine-tuned model, exhibits the largest degradation is notable in light of \cite{wang2025novel}, who find that domain-specific medical LLMs show consistent safety advantages over general-purpose models at full precision. Our results suggest that this domain advantage, where present, is not preserved under aggressive quantization, and may in some cases invert. One possible explanation is that continued pretraining on a narrower, more specialized corpus (PubMed Central, in BioMistral's case) could produce weight distributions with less redundancy than general-purpose pretraining, making the model more sensitive to the per-channel quantization error introduced by GPTQ.

\subsection{Task-Type Sensitivity Has Direct Deployment Implications}

The consistent finding that MedMCQA degrades more than MedQA under INT4, replicated across four of five models, suggests that quantization risk is not only model-dependent but task-dependent within a single model. This is consistent with \cite{wang2025novel}'s broader observation that LLM clinical performance is highly sensitive to scenario complexity and risk stratification, and extends it to the specific context of model compression. 

\subsection{Risk-Stratified Evaluation Reveals Effects Invisible to Aggregate Scores}

The HealthBench risk-weight-5 finding for Qwen2.5-7B ($-26.8\%$ on emergency scenarios versus $-3.0\%$ overall) demonstrates the value of risk-stratified evaluation directly: an aggregate HealthBench score would have substantially understated the clinical risk of INT4 deployment for this model. This finding is directionally consistent with \cite{wang2025novel}'s headline result that LLMs show their largest full-precision performance gaps in high-risk scenarios, and suggests that quantization may compound rather than independently add to this existing high-risk vulnerability. 

\subsection{Safety Alignment Is Independent of Accuracy Robustness}

MedSafetyBench reveals a dimension of quantization risk
invisible to accuracy-only evaluation. Qwen2.5-7B, the most
INT4-robust model on MCQ accuracy benchmarks
($-1.7\%$ MedMCQA), shows the single largest INT4 safety
degradation of any model on MedSafetyBench ($-17.8\%$
refusal rate), indicating that a model's resilience to
quantization-induced accuracy loss provides no guarantee of
resilience to quantization-induced safety degradation. A
practitioner selecting Qwen2.5-7B for INT4 deployment on the
basis of its strong accuracy benchmarks would be unaware
that approximately one in six harmful requests it would
otherwise refuse are instead complied with under compression. 

\subsection{Limitations and Future Work}
\label{sec:limitations}

The HealthBench self-judging paradigm may inflate or deflate scores in a precision-dependent manner if a model's self-evaluation tendencies shift under quantization independently of its generation quality; the anomalous positive INT4 delta for Med42-8B ($+9.3\%$ overall) is a plausible manifestation of this effect and warrants verification using a fixed external judge in future work. The small sample size of the 5 HealthBench subgroup of risk-weight ($n=19$) means that this result should be treated as motivating rather than confirming a risk-amplification effect; larger-scale replication using the full CSEDB benchmark, pending approval of the data access request.

\section{Conclusion}

This study provides a systematic evaluation of the
 effects of post-training quantization on clinical accuracy and safety of LLM across model families, bit-widths, and task types. Three findings were observedfor deployment. First, INT8 GPTQ is a safe default, with minimum degradation across all accuracy benchmarks, while INT4 is not: the degradation of MCQ accuracy ranges from $-19.7\%$ to $0.0\%$ and is
unpredictable from model scale or clinical domain
adaptation -- BioMistral-7B, clinically fine-tuned, shows the worst MCQ degradation in the study, while general-purpose Qwen2.5-7B is among the most robust, and broad knowledge retrieval (MedMCQA) is consistently more fragile than focused reasoning (MedQA). INT4 degradation is larger still on
open-ended and safety-specific evaluation, reaching $-23.0\%$
on HealthBench and $-28.3\%$ on MedSafetyBench, indicating that MCQ benchmarks alone understate INT4's worst-case risk. Second, accuracy robustness and safety-alignment robustness
are independent: Qwen2.5-7B, the most INT4-accuracy-robust model on MedMCQA, shows the largest INT4 safety degradation
on MedSafetyBench ($-17.8\%$ refusal rate), while instruction tuning, not clinical fine-tuning, remains the primary
determinant of baseline safety alignment (base pretrained models refuse harmful requests at $10$--$31\%$ versus
$37$--$75\%$ for instruction-tuned models). Third, mitigating
INT4 degradation is not straightforward: two independent recovery methods, calibration data substitution and QLoRA fine-tuning, produce the same trade-off across four of five models, recovering MedMCQA while further degrading MedQA, suggesting this is a structural property of adapting
quantized models to narrow training distributions rather than a method-specific artifact.

These findings argue against bit-width-based rules of thumb in favor of task-specific and model-specific quantization
safety audits as standard practice. INT8 GPTQ can be broadly recommended; INT4 deployment and any attempt to
recover its losses require empirical validation against every clinical task and evaluation format the model will
 encounter, not just multiple-choice accuracy.

\begin{appendices}
    
\renewcommand{\appendixname}{Appendix}
\section*{Appendix}

\subsection{Appendix A: Clinical Calibration Data Experiment}
\label{app:calibration}

To test whether domain-matched calibration recovers INT4
accuracy loss without full retraining, all five models were re-quantized using MedQA training samples (GPTQ-INT4-Med) in place of WikiText-2 (GPTQ-INT4-Wiki), and evaluated on MedQA and MedMCQA, the benchmarks most degraded under WikiText-2 calibration; Table~\ref{tab:calibration} reports results. Calibration helps base pretrained models lacking clinical QA exposure: Qwen2.5-7B's MedMCQA degradation improves from $-1.7\%$ to $+3.0\%$, and BioMistral-7B's from
$-19.7\%$ to $+10.3\%$. It harms instruction-tuned models,
however: Llama-3.1-8B's MedQA degradation worsens from
$-0.4\%$ to $-49.6\%$, and Med42-8B shows the most extreme
trade-off, MedMCQA more than doubles ($+106.2\%$ over FP16) while MedQA collapses ($-56.4\%$).

\begin{table}[!ht]
\centering
\setlength{\tabcolsep}{1mm}
\fontsize{9pt}{10pt}\selectfont
\begin{tabular}{llccccc}
\toprule
\textbf{Model} & \textbf{BM} & \textbf{FP16} &
\textbf{Wiki} & \textbf{Med} &
\textbf{W$\Delta$} & \textbf{M$\Delta$} \\
\midrule
\multirow{2}{*}{Llama-3.1-8B}
  & MedQA   & 0.533 & 0.531 & 0.269 & $-0.4$ & $-49.6\,$ \\
  & MedMCQA & 0.178 & 0.158 & 0.149 & $-11.2$ & $-16.3\,$ \\
\midrule
\multirow{2}{*}{Qwen2.5-7B}
  & MedQA   & 0.596 & 0.589 & 0.594 & $-1.2$ & $-0.4\,$ \\
  & MedMCQA & 0.297 & 0.292 & 0.306 & $-1.7$ & $+3.0\,$ \\
\midrule
\multirow{2}{*}{BioMistral-7B}
  & MedQA   & 0.410 & 0.413 & 0.391 & $+0.8$ & $-4.6\,$ \\
  & MedMCQA & 0.213 & 0.171 & 0.235 & $-19.7$ & $+10.3\,$ \\
\midrule
\multirow{2}{*}{Meditron-7B}
  & MedQA   & 0.277 & 0.277 & 0.278 & $+0.0$ & $+0.3\,$ \\
  & MedMCQA & 0.590 & 0.590 & 0.527 & $+0.0$ & $-10.7\,$ \\
\midrule
\multirow{2}{*}{Med42-8B}
  & MedQA   & 0.610 & 0.569 & 0.266 & $-6.8$ & $-56.4\,$ \\
  & MedMCQA & 0.272 & 0.260 & 0.561 & $-4.4$ & $+106.2\,$ \\
\bottomrule
\end{tabular}
\caption{Effect of clinical calibration data on INT4 GPTQ accuracy.
\text{BM} = benchmark. \text{Wiki} = GPTQ-INT4-Wiki
(WikiText-2 calibration, standard). \text{Med} = GPTQ-INT4-Med
(128 MedQA training samples as calibration). \text{W$\Delta$} and
\text{M$\Delta$} are $\Delta\%$ from FP16 for Wiki-INT4 and
Med-INT4 respectively.}
\label{tab:calibration}
\end{table}

\subsection{Appendix B: QLoRA Fine-Tuning Recovery}
\label{app:qlora}

As a second approach to mitigating INT4 degradation, we apply
QLoRA fine-tuning~\cite{dettmers2023qlora} to all five
INT4-quantized models using 1,000 MedQA training examples
(LoRA rank 16, alpha 32, 3 epochs, target modules
\texttt{q\_proj}, \texttt{k\_proj}, \texttt{v\_proj},
\texttt{o\_proj}) with results reported in Table~\ref{tab:qlora}. The results replicate the asymmetric task-specific trade-off observed with clinical calibration data in Appendix A: QLoRA fine-tuning recovers MedMCQA performance in four of five
models 
while simultaneously degrading MedQA performance in four of five models. 
That two independent recovery methods, GPTQ calibration data
substitution and QLoRA fine-tuning, produce the same
task-specific trade-off pattern suggests this is a systematic
property of adapting INT4-quantized models toward MedQA training
data, rather than an artifact of either method individually. 

\begin{table}[!ht]
\centering
\setlength{\tabcolsep}{1mm}
\fontsize{9pt}{10pt}\selectfont
\begin{tabular}{llccccc}
\toprule
\textbf{Model} & \textbf{BM} & \textbf{FP16} & \textbf{I4} &
\textbf{QL} & \textbf{I4$\Delta$} & \textbf{QL$\Delta$} \\
\midrule
\multirow{2}{*}{Llama-3.1-8B}
  & MedQA   & 0.533 & 0.531 & 0.508 & $-0.4$ & $-4.9$ \\
  & MedMCQA & 0.178 & 0.158 & 0.360 & $-11.2$ & $+102.2$ \\
\midrule
\multirow{2}{*}{Qwen2.5-7B}
  & MedQA   & 0.596 & 0.589 & 0.539 & $-1.2$ & $-9.6$ \\
  & MedMCQA & 0.297 & 0.292 & 0.353 & $-1.7$ & $+18.9$ \\
\midrule
\multirow{2}{*}{BioMistral-7B}
  & MedQA   & 0.410 & 0.413 & 0.312 & $+0.8$ & $-23.9$ \\
  & MedMCQA & 0.213 & 0.171 & 0.305 & $-19.7$ & $+43.2$ \\
\midrule
\multirow{2}{*}{Meditron-7B}
  & MedQA   & 0.277 & 0.277 & 0.277 & $+0.0$ & $-0.3$ \\
  & MedMCQA & 0.590 & 0.590 & 0.563 & $+0.0$ & $-4.6$ \\
\midrule
\multirow{2}{*}{Med42-8B}
  & MedQA   & 0.610 & 0.569 & 0.452 & $-6.8$ & $-26.0$ \\
  & MedMCQA & 0.272 & 0.260 & 0.275 & $-4.4$ & $+1.1$ \\
\bottomrule
\end{tabular}
\caption{QLoRA fine-tuning recovery results. LoRA rank 16, 1,000
MedQA training examples, 3 epochs. \text{BM} = benchmark.
\text{I4} = INT4 (GPTQ, pre-recovery). \text{QL} = QLoRA
fine-tuned. \text{I4$\Delta$} and \text{QL$\Delta$} are
$\Delta\%$ from FP16 for INT4 and QLoRA respectively. Recovery =
QLoRA $-$ INT4.}
\label{tab:qlora}
\end{table}
\end{appendices}

\bibliography{aaai2027}

\end{document}